\documentclass[a4paper,fleqn]{cas-dc}
\usepackage{comment}
\usepackage{color}
\usepackage{algorithm}
\usepackage{graphicx}
\usepackage{algpseudocode}
\usepackage{amsmath, amssymb}
\usepackage[justification=centering]{caption}
\usepackage{multirow}
\usepackage{float}
\usepackage{lipsum}
\usepackage{bbm}
\usepackage{bm}
\usepackage{subfigure}
\usepackage{mathtools}
\usepackage{comment}
\usepackage{arydshln}
\usepackage[normalem]{ulem}
\useunder{\uline}{\ul}{}

\newcommand{\qin}[1]{{\textcolor{black}{#1}}}

\usepackage[numbers]{natbib}
\def\tsc#1{\csdef{#1}{\textsc{\lowercase{#1}}\xspace}}
\tsc{WGM}
\tsc{QE}

\begin{document}
\let\WriteBookmarks\relax
\def\floatpagepagefraction{1}
\def\textpagefraction{.001}
\shorttitle{Multi-Scale Distillation for RGB-D Anomaly Detection on the PD-REAL Dataset}
\shortauthors{J. Qin et al.}
\title [mode = title]{Multi-Scale Distillation for RGB-D Anomaly Detection on the PD-REAL Dataset}

\author[1]{Jianjian Qin}

\author[2]{Chao Zhang}

\author[3]{Chunzhi Gu}

\author[4]{Zi Wang}

\author[4]{Jun Yu}

\author[2]{Yijin Wei}

\author[1]{Hui Xiao}

\author[1]{Xin Yu}

\affiliation[1]{organization={Ningbotech University},
            city={Ningbo},
            postcode={315100},
            state={Zhejiang},
            country={China}}

\affiliation[2]{organization={University of Toyama},
            city={Toyama-shi},
            postcode={930-8555},
            state={Toyama},
            country={Japan}}

\affiliation[3]{organization={University of Fukui},
            city={Fukui-shi},
            postcode={910-8507},
            state={Fukui},
            country={Japan}}

\affiliation[4]{organization={Niigata University},
            city={Niigata-shi},
            postcode={950-2181},
            state={Niigata},
            country={Japan}}
\begin{abstract}
We present PD-REAL, a novel large-scale dataset for unsupervised anomaly detection (AD) in the 3D domain. It is motivated by the fact that 2D-only representations in the AD task may fail to capture the geometric structures of anomalies due to uncertainty in lighting conditions or shooting angles. PD-REAL consists entirely of Play-Doh models for 15 object categories and focuses on the analysis of potential benefits from 3D information in a controlled environment. Specifically, objects are first created with six types of anomalies, such as \textit{dent}, \textit{crack}, or \textit{perforation}, and then photographed under different lighting conditions to mimic real-world inspection scenarios. To demonstrate the usefulness of 3D information, we use a commercially available RealSense camera to capture RGB and depth images. Compared to the existing 3D dataset for AD tasks, the data acquisition of PD-REAL is significantly cheaper, easily scalable, and easier to control variables. Furthermore, we introduce a multi-scale teacher--student framework with hierarchical distillation for multimodal anomaly detection. This architecture overcomes the inherent limitation of single-scale distillation approaches, which often struggle to reconcile global context with local features. Leveraging multi-level guidance from the teacher network, the student network can effectively capture richer features for anomaly detection. Extensive evaluations with our method and state-of-the-art AD algorithms on our dataset qualitatively and quantitatively demonstrate the higher detection accuracy of our method. Our dataset can be downloaded from \url{https://github.com/Andy-cs008/PD-REAL}
\end{abstract}

\begin{keywords}
Anomaly Detection \sep Multi-Scale-Distillation  \sep Surface Inspection  \sep Depth Image
\end{keywords}

\maketitle

\section{Introduction}
\label{sec:introduction}

Anomaly detection (AD), which plays a vital role in various research fields, such as industrial inspection \cite{gudovskiy2022cflow,bergmann2020uninformed,roth2022towards,wang2021student_teacher, wang2022improved} and medical image processing \cite{cai2023dual, jia2022attention, schlegl2019f}, has received increasing attention recently.
It aims to accurately detect and segment the anomalies in samples for visual surface inspection. So far, because of the clear visibility and localization of anomaly patterns in real-world applications, the development of AD techniques has progressed mainly for 2D images. 

\begin{figure} \centering    
\subfigure[RGB] {
 \label{fig:a}     
\includegraphics[width=0.28\columnwidth]{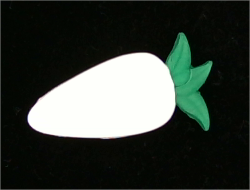} }     
\subfigure[3D point cloud] { 
\label{fig:b}     
\includegraphics[width=0.28\columnwidth]{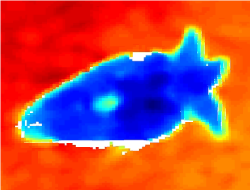}     
}    
\subfigure[Ground truth] { 
\label{fig:c}     
\includegraphics[width=0.28\columnwidth]{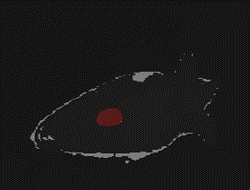}     
}  
\caption{An example of the anomalous sample with the anomaly type \textit{dent}. 
The anomaly is clearly visible in the 3D point cloud (b) compared to the 2D RGB image (a).}

\label{fig:1}     
\end{figure}

One of the key challenges of AD for images arises from the fact that 2D representations can easily encounter bottlenecks in capturing complete geometric structures. Fig. \ref{fig:1} illustrates an example of this problem. It is usually the case that the acquired image (Fig. \ref{fig:1}(a)) does not provide a clear visual indication of the anomaly locations. Due to the intrinsic color and reflected light of the object, the appearance of the target may vary significantly from different shooting angles. This problem can be worsened by varying scene lighting conditions and potential over/under exposure during acquisition.


Compared to 2D image, 3D data generally provides a more powerful and comprehensive presentation for the real world \cite{saleh2013object, chang2015shapenet,armeni20163d,yadav2023habitat,huang2018apolloscape,xiao2021pandaset}. Therefore, a straightforward solution to the above challenge would be to use 3D data representations in combination with 2D images. For example, the subtle anomaly in Fig. \ref{fig:1}(a) can be clearly observed using a 3D point cloud (Fig. \ref{fig:1}(b)).
In view of this, we argue that explicitly exploiting 3D information for AD tasks is more reasonable. However, due to the previous lack of suitable datasets, AD in the 3D domain was initially explored less extensively. In recent years, several public datasets have become available for 3D AD. Among the early and representative ones are MVTec 3D-AD \cite{bergmann2021mvtec} and Eyecandies \cite{bonfiglioli2022eyecandies}. For MVTec 3D-AD, it involves the usage of highly expensive industrial sensors for data acquisition. It should be noted that in real-world AD applications, more samples may be frequently and extensively required to meet the increasing industrial demands, and such a costly manner of data acquisition and storage may hinder dataset expansion. As for Eyecandies, all the data is collected in a virtual environment to simulate large defect variations. However, virtual anomalies still show inherent differences to the real-world ones, which can cause the issue of domain bias and reflect poorly the real geometric features.

Motivated by the above analysis, we present a novel dataset named PD-REAL for unsupervised AD in the 3D domain. All of the samples in PD-REAL are handmade with Play-Doh. In particular, we first make 15 object categories by referring to the guide, such as Play-Doh molds. Then, to mimic the inspection scenarios in practice, we manually create six types of anomalies. The Intel RealSense Depth Camera D405 (RealSense 405) is used for data acquisition. It is elevated and placed statically above the object to record RGB and depth images per object from the same angle. We then produce the whole 3D information (i.e., point cloud) for each object using the corresponding image pair with the camera intrinsics. For each category, we provide a training, validation, and test set, respectively.  In total, PD-REAL contains more than 3,500 RGB and depth image pairs. Moreover, because of the editable nature of Play-Doh, it is far less burdensome to expand and adjust the shape of the objects to include more different types of anomalies as needed. To fully assess PD-REAL, we benchmark state-of-the-art AD methods on it and experimentally analyze the effect of exploiting  3D information for AD.
\begin{figure*}[tb]
     \centering
     \includegraphics[width=1\linewidth]{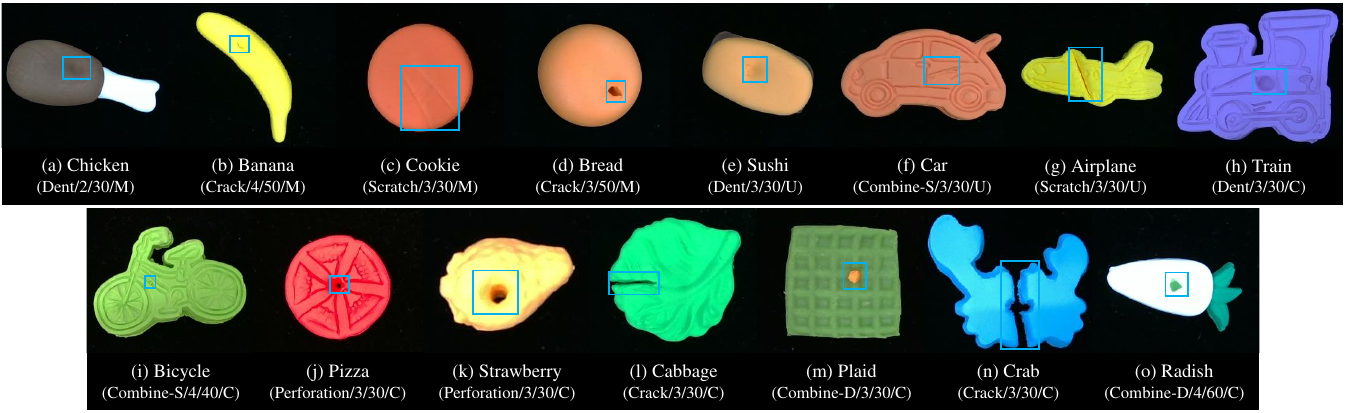}
     \caption{Example images from the proposed PD-REAL dataset. Under each image, the first row indicates the category, and the second row indicates the anomaly type highlighted with the blue box/category-wise number of anomaly types/category-wise number of total anomalous samples/lighting condition. The lighting conditions are shortened in C: controlled, U: uncontrolled, and M: mixed for brevity.
}
     \label{fig:2}
\end{figure*}


In summary, this paper consists of the following main contributions: 
\begin{itemize}

\item We propose a novel large-scale 3D AD dataset PD-REAL with Play-Doh made samples. It covers 15 object categories and six anomaly types recorded under multiple scene illuminations. 

\item An efficient sample collection pipeline is developed using an Intel RealSense camera for data acquisition, offering a substantially lower-cost and more easily extensible alternative to existing 3D AD datasets.

\item\qin{We introduce a multi-scale teacher--student framework with hierarchical distillation for anomaly detection. By aggregating local, intermediate, and global features, the student learns richer priors and improves detection accuracy.}

\item \qin{We benchmark our framework against state-of-the-art anomaly detection methods. Results show superior performance over competitive baselines, both qualitatively and quantitatively.}

\end{itemize}

\section{Related Work}
\label{sec:Related Work}
In this section, we review prior studies on AD datasets and methods. We first summarize 2D and 3D AD datasets, then review 2D AD methods, and finally discuss representative 3D AD approaches.

\subsection{2D Anomaly Detection Datasets}

Most existing AD datasets are developed for 2D images. MVTec AD \cite{bergmann2019mvtec} is a well-known benchmark and the first AD dataset with pixel-level ground-truth annotations. It contains 5,354 high-resolution color images, including five texture categories and ten object categories, with over 70 anomaly types.
Other datasets have also been introduced for different AD tasks. The ShanghaiTech Campus dataset \cite{liu2018future} is designed for anomalous event detection in surveillance videos and mainly targets non-industrial scenarios. The dataset proposed by Huang et al. \cite{huang2020surface} focuses on magnetic-tile surface defect detection. The NanoTWICE dataset \cite{carrera2016defect} includes high-resolution images for nanofibrous materials; however, its sample size is limited for training deep learning models.
DAGM \cite{wieler2007weakly} is a dataset for textured-surface defect detection, containing images of various surfaces with artificially generated defects. Silvestre et al. \cite{silvestre2019public} released a public fabric defect dataset with 245 images across seven fabric types. More recently, Zou et al. \cite{zou2022spot} proposed Visual Anomaly (VisA) for both one-class and two-class AD tasks, containing 10,821 high-resolution color images that cover diverse industrial objects.

\subsection{3D Anomaly Detection Datasets}
3D AD remains less explored than 2D AD, largely due to the limited availability of suitable public datasets. To the best of our knowledge, two representative public datasets are MVTec 3D-AD \cite{bergmann2021mvtec} and Eyecandies \cite{bonfiglioli2022eyecandies}. MVTec 3D-AD \cite{bergmann2021mvtec} contains 4,147 high-precision 3D point clouds from ten real-world object categories. It includes 41 defect types, with 2,656 normal training samples and 249 normal plus 948 anomalous test samples. The data are acquired using an expensive industrial sensor (Zivid One Medium). Bonfiglioli et al. \cite{bonfiglioli2022eyecandies} proposed Eyecandies, a synthetic 3D dataset for unsupervised multimodal AD and localization. It provides 90,000 photo-realistic renderings of synthetic objects in virtual scenes, covering ten categories. For each sample, six lighting conditions are provided together with ground-truth depth and normal maps. However, a domain gap between synthetic and real defects still limits its generalization to real-world AD tasks. By contrast, PD-REAL was first introduced in our previous work \cite{qin2023pd} and was collected from real objects. Built with an Intel RealSense camera and Play-Doh samples, PD-REAL substantially reduces collection cost and can be conveniently extended thanks to the malleability of Play-Doh.

\subsection{2D Anomaly Detection Methods}
Benefiting from abundant 2D AD datasets, mainstream AD methods have primarily focused on the 2D domain \cite{roth2022towards,cai2023itran,wan2023anomaly,cohen2020sub,defard2021padim, salehi2021multiresolution,yang2023memseg}. PatchCore \cite{roth2022towards} is a representative 2D AD method that uses a memory bank of patch-level normal features to distinguish anomalous regions. Salehi et al. \cite{salehi2021multiresolution} adopted a knowledge distillation scheme by modeling discrepancies between teacher and student intermediate features for AD.
Different from the unsupervised settings of \cite{roth2022towards,salehi2021multiresolution}, Yang et al. \cite{yang2023memseg} proposed an end-to-end semi-supervised framework that fuses memory information with multi-scale image features for improved localization. Nevertheless, these methods are designed for 2D AD tasks. Without explicit depth cues, 2D representations cannot fully capture object geometry. Therefore, directly applying these methods to 3D scenarios is insufficient to model the unordered and noisier characteristics of 3D data.

\subsection{3D Anomaly Detection Methods}
To the best of our knowledge, 3D AD is still less explored than 2D AD. Existing methods can be grouped into two settings: (a) surface-defect AD and (b) general-object AD. For (a), Bergmann et al. \cite{bergmann2023anomaly} introduced a teacher--student framework with local receptive-field reconstruction, and Rudolph et al. \cite{rudolph2023asymmetric} further proposed an asymmetric teacher--student design with a bijective normalizing flow to reduce over-generalization. Horwitz et al. \cite{horwitz2022empirical} conducted an empirical study that extracts 3D point-cloud features using Fast Point Feature Histograms (FPFH), combines them with RGB features from WideResNet50 \cite{zagoruyko2016wide}, and feeds the fused features into PatchCore. Wang et al. \cite{wang2023multimodal} proposed multimodal hybrid feature fusion with patch-wise contrastive learning, while Chen et al. \cite{chen2023easynet} introduced attention mechanisms for multimodal 3D industrial AD. For (b), the goal is to determine whether an input 3D object belongs to the user-defined normal or anomalous class. Floris et al. \cite{floris2022composite} used composite layers to model point spatial structure, and Masuda et al. \cite{masuda2021toward} adopted a variational autoencoder and detected anomalies via reconstruction error. However, reconstruction-based methods usually require many training samples and higher computational cost. To address this limitation, Qin et al. \cite{qin2023teacher} proposed a teacher--student framework that works with only a few normal samples, where multi-scale loss is used to transfer knowledge from teacher to student for more accurate feature learning.

\section{PD-REAL}
\label{sec:PD-REAL}

\subsection{Dataset Description}

PD-REAL consists of over 3,500 high-resolution ($640\times 480$) RGB images and the corresponding depth images and point clouds. Fig. \ref{fig:2} illustrates some example RGB image samples per category of PD-REAL. The details of the PD-REAL samples are described below:\

 \noindent	\textbf{Categories.} PD-REAL contains 15 object categories, which can be grouped as: Food (chicken, cookie, bread, sushi, pizza), Vegetables (cabbage, radish), Fruits (banana, strawberry), and Toys (car, airplane, train, bicycle, plaid, crab).\par
 \noindent	\textbf{Images.} For each sample, we record two images: RGB and depth images. The image pair is then utilized to generate the corresponding 3D point cloud.\par
\begin{figure}[]
     \centering
     \includegraphics[width=1\linewidth]{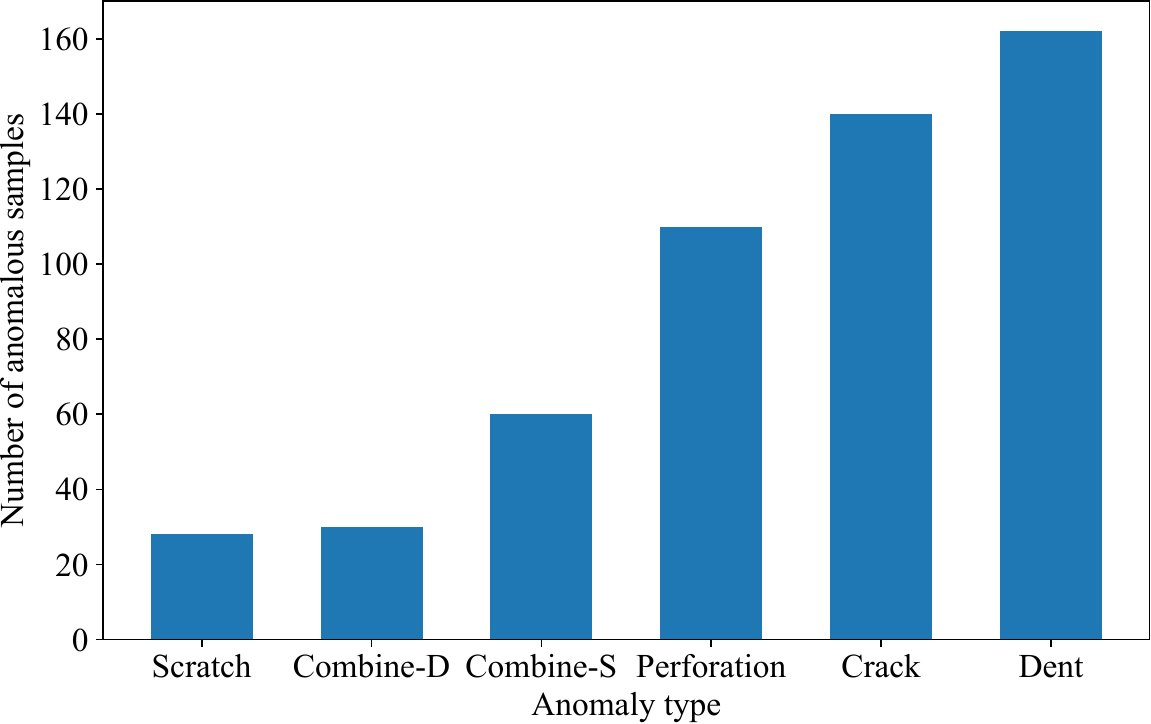}
     \caption{{Distribution of anomalous samples in PD-REAL across the six anomaly types and 15 object categories. The histogram summarizes the number of anomalous samples per category.}} 
     \label{fig:3}
\end{figure}

 \noindent	\textbf{Sample distribution.} For each category, we provide a training, validation, and test set, respectively. To fully address the task of unsupervised AD, only normal samples are included in the training and validation sets. The test sets contain both normal and anomalous samples. The anomalous samples are manually created with six types of anomalies: \textit{dent}, \textit{crack}, \textit{perforation}, \textit{scratch}, \textit{combine-S}, and \textit{combine-D}. Specifically, for \textit{combine-S} and \textit{combine-D}, we combine a foreign object with the same or different colors to the original sample. Please refer to Fig. \ref{fig:3} for a detailed statistical distribution of the anomalous samples.\par
 
\noindent	\textbf{Lighting conditions.} We set three types of scene lighting conditions during shooting: controlled light (C), uncontrolled light (U), and mixed light (M). Each category follows one of these lighting conditions. Specifically, for C, we shoot the samples under an indoor configuration with a fixed position light source, while for U, we expose the samples outdoors to allow multiple illumination from arbitrary angles. M denotes a mixture of U and C for recording.

\subsection{Data Collection}
Previous 3D datasets \cite{bergmann2021mvtec} for AD depend on industry-level sensors for data collection. Although this allows a higher image resolution, it can cause the extension of the dataset to be highly expensive. To provide a practical solution to 3D collection, we design an efficient recording pipeline that enables easy extension. Specifically, we use the RealSense D405\footnote{https://www.intelrealsense.com/depth-camera-d405/} to collect all the data. This is a short-range stereo camera with sub-millimetre accuracy that can capture the RGB and depth images simultaneously. We elevate it so that it can capture the whole object from above. Before shooting, we use a leveller to ensure that the background of the object is parallel to the camera plane to reduce the depth error.



\noindent	\textbf{Sample generation.}  In PD-REAL, all samples are handmade with Play-Doh. Contrary to the conceptual simpleness, it leads to great flexibility in easily constructing diverse types of objects due to the scalable nature of Play-Doh.
\par
\noindent	\textbf{Anomaly annotation.} For each anomalous sample, we manually annotate the pixel-wise anomalous region for ground truth, using LabelMe \footnote{http://labelme.csail.mit.edu/Release3.0/}.\par
\noindent	\textbf{Point cloud generation.} Given the acquired RGB $I_{RGB}$ and depth image $I_{d}$ pair, we generate the corresponding point cloud to represent the full 3D information. To bridge the gap between the two coordinate systems of the pixel and the world, we relate any pixel located at $(u, v)$ in $I_{RGB}$ to its 3D coordinate $(x_w, y_w, z_w)$ using the following transformation:
\begin{equation}
\label{eq:eq1}
{\small
z_{c}
\begin{pmatrix}
u\\
v\\
1
\end{pmatrix}
=
\begin{bmatrix}
f/dx & 0 & u_0\\
0 & f/dy & v_0\\
0 & 0 & 1
\end{bmatrix}
\begin{bmatrix}
R & T
\end{bmatrix}
\begin{pmatrix}
x_{w}\\
y_{w}\\
z_{w}\\
1
\end{pmatrix}
,}
\end{equation}
where $z_c$ is the $z$-axis value of the camera coordinates, i.e., the distance from the object to the camera. $f$ is the focal length of the camera, and $dx,dy$ are the length and width of an image element. $u_0,v_0$ are coordinates of the origin of the image coordinate system under the pixel coordinate system. $R,T$ are the $3 \times 3$ rotation matrix and the $3 \times 1$ translation matrix of the external reference matrix. Since the camera coordinate system is identical to the world coordinate system, the depth of an object is shared in both systems (i.e., $z_c = z_w$). Thus, $R$ and $T$ can be simplified as an identity matrix and a zero matrix, respectively. The world coordinates for the point cloud can then be solved using Eq. \ref{eq:eq1}.

\begin{figure*}[h]
     \centering
     \includegraphics[width=1\linewidth]{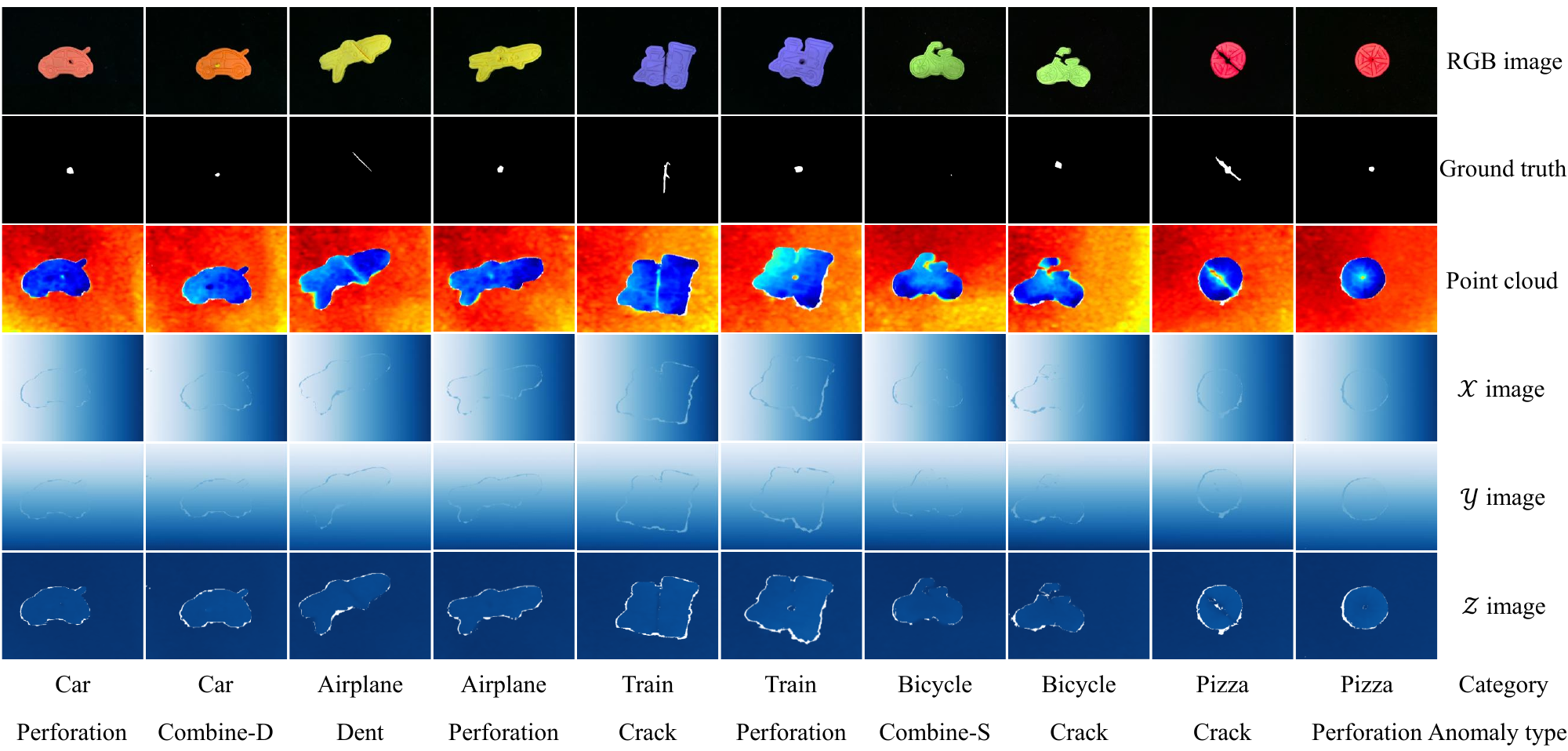}
     \caption{PD-REAL visualizations for car, airplane, train, bicycle, and pizza. For each category, two anomalous samples are shown. From top to bottom: RGB image,  ground-truth , 3D point cloud, and 3D coordinate encodings along the $x$-, $y$-, and $z$-axes.} 
     \label{fig:4}
\end{figure*}

\begin{figure*}[h]
     \centering
     \includegraphics[width=1\linewidth]{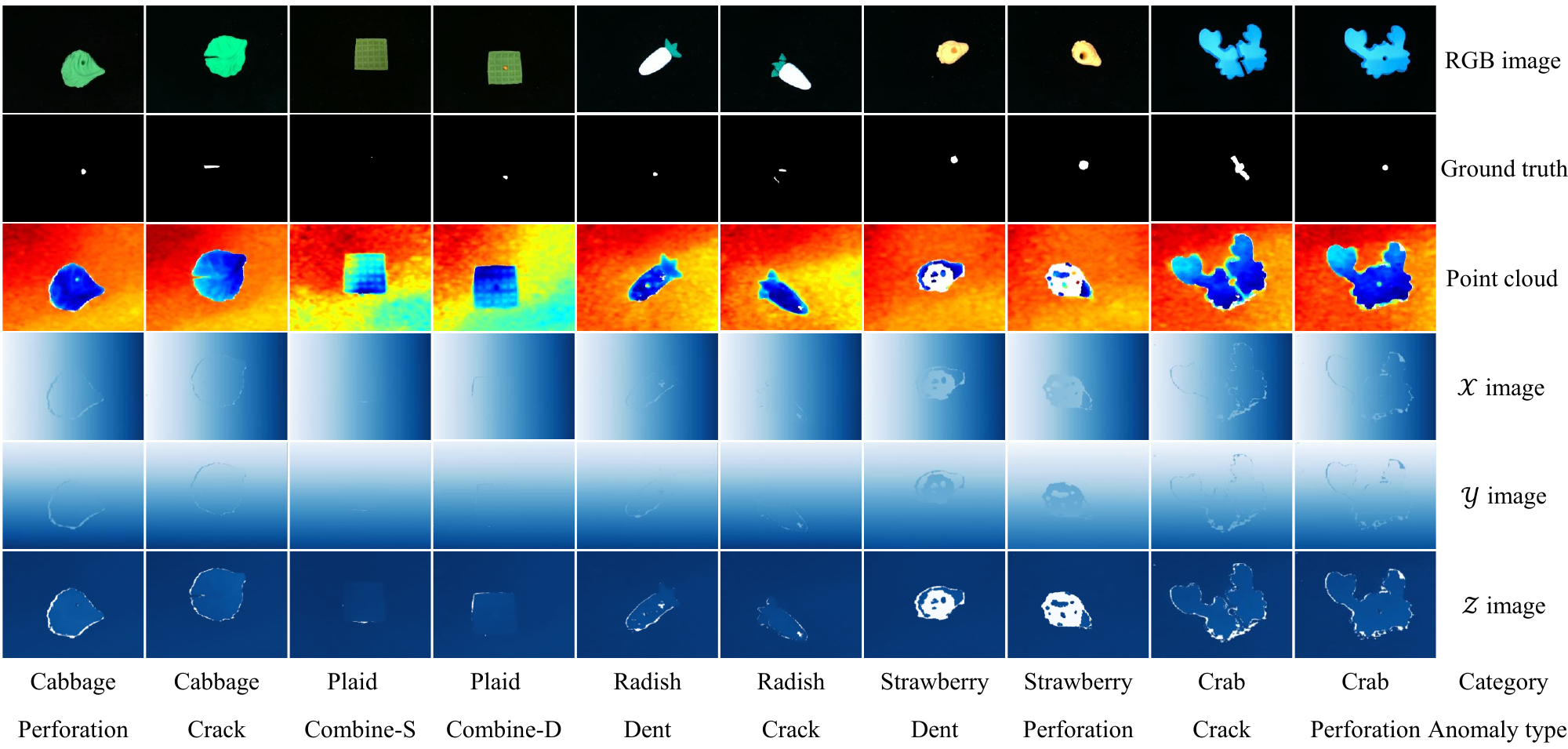}
     \caption{PD-REAL visualizations for cabbage, plaid, radish, strawberry, and crab. For each category, two anomalous samples are shown. From top to bottom: RGB image, ground-truth, 3D point cloud, and 3D coordinate encodings along the $x$-, $y$-, and $z$-axes.} 
     \label{fig:6}
\end{figure*}

More visualizations of our dataset are depicted in Fig. \ref{fig:4} $\sim$ \ref{fig:6}. For each category, we show the visual representations of two example anomalous samples. In Fig. \ref{fig:4} $\sim$ \ref{fig:6}, we present the RGB image, pixel-level ground-truth anomaly mask, and the 3D point cloud in three rows on the top. The other three rows on the bottom visualize the 3D coordinates encoding of objects along the $x-$, $y-$, and $z$-axis.

\begin{figure*}[h]
     \centering
     \includegraphics[width=1\linewidth]{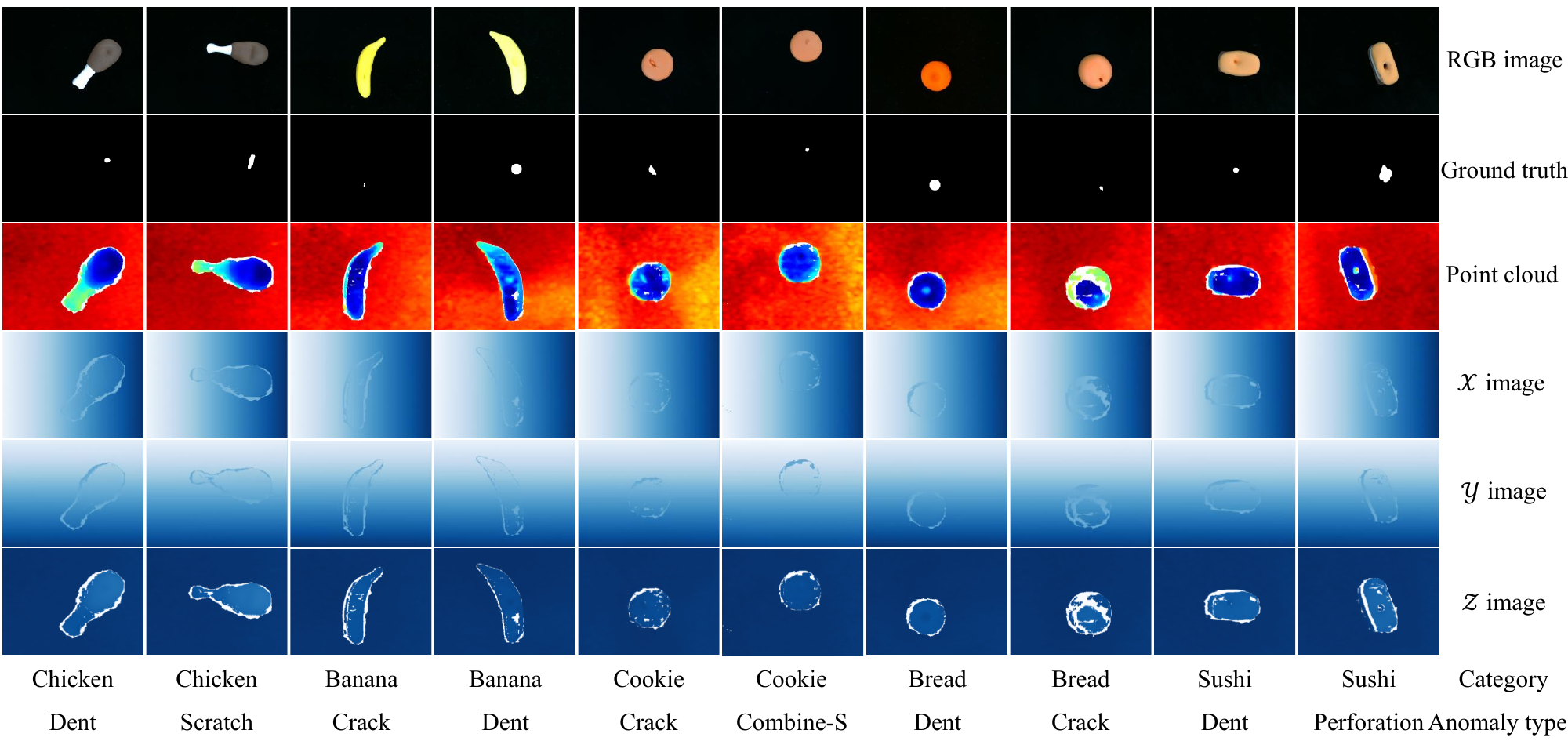}
     \caption{PD-REAL visualizations for chicken, banana, cookie, bread, and sushi. For each category, two anomalous samples are shown. From top to bottom: RGB image, ground-truth, 3D point cloud, and 3D coordinate encodings along the $x$-, $y$-, and $z$-axes.} 
     \label{fig:5}
\end{figure*}

\begin{figure*}[tb]
     \centering
     \includegraphics[width=1\linewidth]{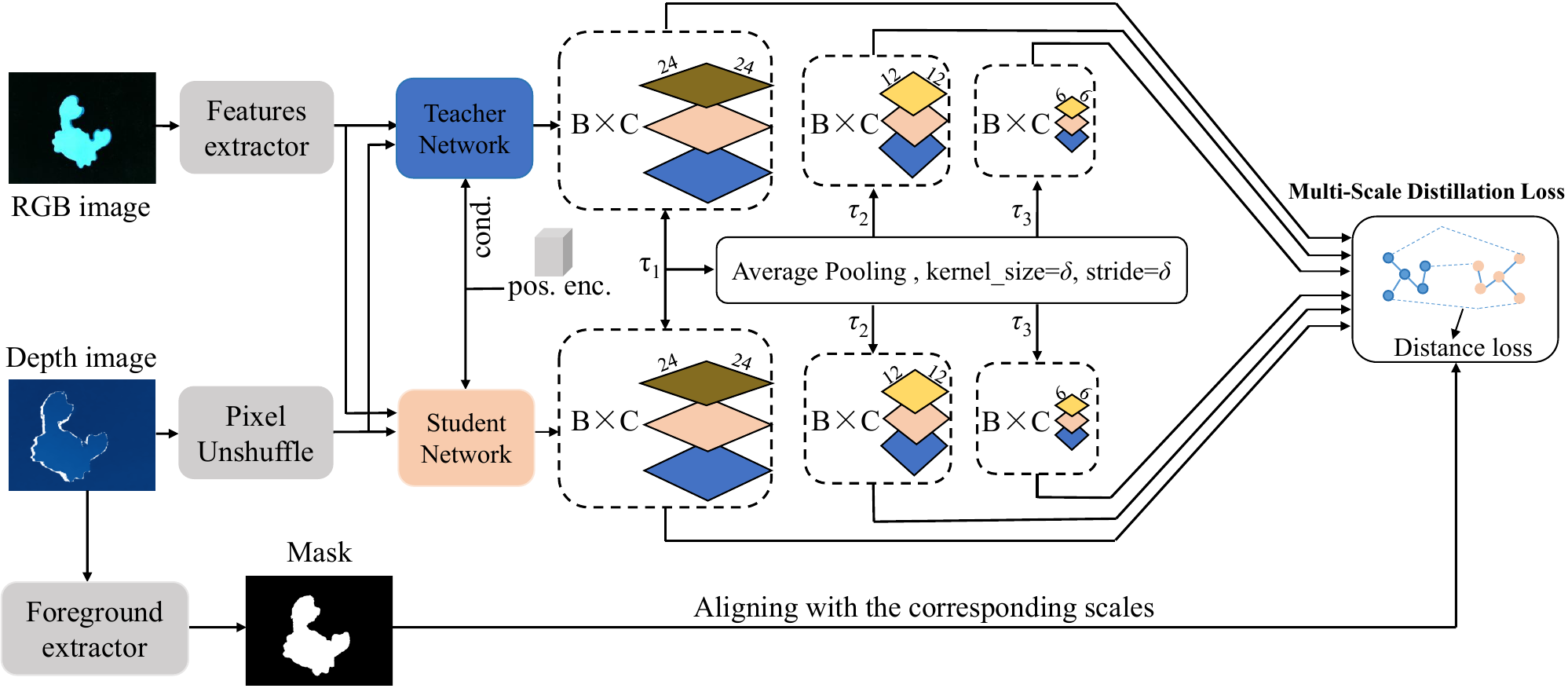}
     \caption{{Overview of multi-scale distillation framework for 3D anomaly detection. In the training stage, the student network is optimized on normal samples to align its multi-scale features with the teacher's outputs. During the testing stage, the anomaly score is derived from the $\ell_2$ distance between the teacher and student representations.}} 
     \label{fig:7}
\end{figure*}

\section{Multi-Scale Distillation Framework}
Drawing inspiration from the efficacy of multi-scale distillation mechanisms in capturing cross-scale features for anomaly detection \cite{iqbal2024multi, qin2023teacher}, \qin{building upon \cite{rudolph2023asymmetric}, we introduce a multi-scale distillation mechanism to overcome the inherent constraints of single-scale paradigms. Traditional single-scale approaches often fail to reconcile global features with fine-grained local features. Our framework bridges this gap by facilitating hierarchical feature alignment across diverse receptive fields. By integrating multi-scale feature extraction with a composite loss formulation, the proposed framework encourages a more balanced representation across the semantic level, thereby yielding greater improvement in multimodal anomaly detection accuracy. Specifically, the teacher network employs a normalizing flow to learn a bijective mapping from the training distribution ${p_X}$ to a standard normal distribution $\mathcal{N}(0,1)$. The student network is optimized through multi-scale knowledge distillation to mimic the output of the teacher network, where the objective is to minimize the $\ell_2$  distance between the output of the teacher and the student network. Fig. \ref{fig:7} shows the distillation process of our multi-scale teacher--student network. }\par
\noindent	\textbf{Teacher-student architecture.} 
\qin{The teacher network employs a conditional normalizing flow, whereas the student model consists of a standard convolutional neural network. The feature extractor is the layer output of EfficientNet-B5 \cite{tan2019efficientnet} pretrained on ImageNet \cite{deng2009imagenet}. For the RGB modality, the teacher--student network utilizes the feature extracted by the feature extractor rather than the raw pixel data. For the point cloud, the $x$- and $y$- coordinates are discarded to focus on the depth information along the $z$-axis, which is subsequently down-sampled through pixel unshuffle to ensure spatial alignment with the RGB features for concatenation. Positional encoding is obtained by employing a sinusoidal positional encoding scheme \cite{vaswani2017attention} in the spatial dimensions of the input feature maps. In the teacher network, the positional encoding serves as a conditional input to control the computation of transformation parameters,  while in the student network, the positional encoding is directly concatenated with  the input feature representations.} \par

{\noindent	\textbf{Multi-scale feature aggregation.} During the training process, we adopt a set of $N=3$ scales, denoted by $\tau \in \mathcal{T}=\{\tau_1,\tau_2,\dots,\tau_N\}$. The finest scale ($\tau_1$) corresponds to the original output features of the teacher and student networks. Progressively coarser representations are obtained by average pooling applied to the original output. The kernel size and stride are determined by the scale factor $\delta$, specifically configured as $\tau_2$ for $\delta=2$ and $\tau_3$ for $\delta=4$. The resulting multi-scale architecture is \qin{illustrated} in Fig. \ref{fig:7}. This hierarchical aggregation captures fine-grained, intermediate, and global representations, ensuring the student model effectively learns multi-resolution feature embeddings.  \par
{\noindent	\textbf{Spatial supervision with multi-scale masks.}  Mask is derived from the depth map by computing the distance of each pixel from the modeled background plane. To spatially supervise the distillation, the binary mask demarcates the specific regions designated for loss computation. Throughout the training phase, this mask is dynamically adjusted by employing scale-specific pooling and re-binarization to maintain spatial alignment with the multi-resolution feature maps.}\par
{\noindent \textbf{Multi-scale loss function.} The total loss is then computed as the mean loss across these distinct scales. During the testing process, following \cite{qin2023teacher}, the distillation hierarchy is bypassed, and the evaluation is conducted using only the original raw outputs from both the teacher and student networks. To optimize the student network, first, we designate the student network as $F_\text{s}$, which is trained to emulate the feature representation produced by a frozen teacher network $F_\text{t}$. The input $x$ comprises a concatenation of image features, 3D information, and positional encoding. To ensure compatibility between the two representation spaces, the output dimensionality of $F_\text{s}$ is explicitly aligned with that of $F_\text{t}$.\par
During the training process, at scale $\tau\in\mathcal{T}$, the pixel-wise distillation loss is defined as the masked squared $\ell_2$ distance between the student and teacher feature vectors:

\begin{equation}
\label{eq:ms-loss}
    \mathcal{L}^{(\tau)}_{ij} = m^{(\tau)}_{ij} \cdot
    \left\| F^{(\tau)}_\text{t}(x)_{ij} - F^{(\tau)}_\text{s}(x)_{ij} \right\|_2^2,
\end{equation}
where  $F^{(\tau)}_\text{s}(x)_{ij}$ and $F^{(\tau)}_\text{t}(x)_{ij}$ denote the output feature vectors of the student and teacher networks at spatial location $(i,j)$ under scale $\tau$, respectively, and $m^{(\tau)}_{ij} \in \{0,1\}$ is the binary foreground mask at $\tau$ scale that suppresses background contributions. The overall training objective is obtained by averaging over all spatial locations and scales:

\begin{equation}
  \label{eq:3}
    \mathcal{L}_{train} = \mathbb{E}_{\tau}\!\left[\mathbb{E}_{i,j}
    \!\left[\mathcal{L}^{(\tau)}_{ij}\right]\right],
\end{equation}
where $\mathbb{E}[\cdot]$ denotes the empirical mean over the subscripted indices.
The model is optimized by minimizing $\mathcal{L}_{train}$ across $N$ scales and
all training samples.\par
During the testing process, we only use the anomaly score at the first scale ($\tau_1$). Specifically, pixel-wise anomaly scores are computed according to Eq. \ref{eq:ms-loss}, yielding a spatially-resolved map for localization. Since anomalous regions in our dataset are predominantly small and localized, for image-level detection, we aggregate pixel-wise anomaly scores by taking the spatial maximum over all pixel locations. By isolating the highest pixel response, the model maintains high sensitivity to tiny anomalies. 

\section{Experiments}
\label{sec:Experiments}

\subsection{Evaluation Metrics}
Following \cite{bergmann2021mvtec,bergmann2019mvtec}, we adopt the
per-region overlap (PRO) metric to report the pixel-level AD (i.e., anomaly localization) performance. Given the anomaly score produced by the algorithms, we binarize it according to a predetermined threshold $\lambda$ to obtain the final prediction $P$. The PRO can then be computed by measuring the average component-wise relative overlap with the ground truth:
\begin{equation}
\label{eq:eq2}
\text{PRO}= \frac{1}{K}\sum_{k=1}^{K}\frac{\lvert{P\cap C_{k}}\rvert}{\lvert C_{k} \rvert},
\end{equation}
where $K$ is the total number of ground truth components and $C_{k}$ is the $k$th ground truth. By increasingly varying $\lambda$ until an average per pixel false positive rate of 30\% is reached for the entire dataset (as suggested in \cite{bergmann2021mvtec}), we plot the PRO curve and integrate the area under such a curve to measure the final PRO value (AUPRO). We also use the area under the receiver operating characteristic curve (AUROC) \cite{bradley1997use} to measure the accuracy of anomaly detection at the image level. The receiver operating characteristic (ROC) curve reflects the performance of the binary classification model under various thresholds. The AUROC then computes the whole area under the ROC curve, and a larger AUROC indicates a higher image-level detection performance.

\subsection{Compared Methods}

We evaluate two state-of-the-art 3D AD techniques: AST \cite{rudolph2023asymmetric} and M3DM \cite{wang2023multimodal} 
on our dataset for assessment. AST \cite{rudolph2023asymmetric} utilizes an asymmetric teacher--student network with a bijective normalizing flow-based teacher architecture to mitigate the over-generalization in distillation. M3DM \cite{wang2023multimodal} performs unsupervised hybrid feature fusion on multimodal data for 3D AD. We train AST and M3DM on our datasets following the provided parameter settings. We also carefully tune the parameters during data preprocessing in background removal so that the object area can be fully retained to facilitate model learning.

Besides the above two 3D AD techniques, we further train a state-of-the-art 2D AD method $-$ PatchCore \cite{roth2022towards} on our dataset and provide analysis. In particular, we follow \cite{horwitz2022empirical} by adopting three PatchCore variants: RGB iNET, Depth iNET, and RGB+FPFH. For RGB iNET and Depth iNET, we train PatchCore only on RGB and Depth images, respectively. For RGB+FPFH, we first use FPFH \cite{rusu2009fast} to extract features from the 3D point cloud. These features are then concatenated with RGB images to feed into PatchCore for further learning. Note that the image features for PatchCore are obtained by using WideResNet50 \cite{zagoruyko2016wide} pre-trained on the ImageNet dataset \cite{deng2009imagenet} for all three scenarios. For all of the above PatchCore-based methods, the input point cloud and color images are down-sampled with nearest-neighbor interpolation and bicubic interpolation, respectively, following \cite{horwitz2022empirical}. The depth map directly utilizes the z channel of point cloud. Finally, we evaluated our dataset against UniNet \cite{wei2025uninet}, a state-of-the-art unified anomaly detection framework. UniNet enhances cross-domain generalization through domain-related feature selection and contrastive learning. Our experiments strictly followed the original training and testing protocols. Specifically, consistent with UniNet's approach to the MVTec 3D-AD dataset, we trained the model on our dataset using exclusively RGB information.

\begin{table*}[h]
\centering
\caption{{Object category-wise AUPRO on PD-REAL. Results are reported per category, with Mean denoting the average AUPRO across all categories. Best and second-best results are highlighted in bold and underlined, respectively.}}
\label{tab:tab1}
\scalebox{0.8}{
\begin{tabular}{ccccccccc}
\toprule
Method     & Chicken & Banana & Cookie    & Bread   & Sushi & Car   & Airplane & Train \\ \toprule
RGB iNet     & 0.927   & 0.943  & 0.743  & \textbf{0.981}   & 0.980 & 0.886 & 0.945    & 0.954 \\
Depth iNet   & 0.687   & 0.791  & 0.805  & 0.882   & 0.886 & 0.507 & 0.568    & 0.554 \\
RGB+FPFH     & 0.954   & 0.943  & 0.868  & \underline{0.979}   & \underline{0.983} & 0.876 & 0.959    & \underline{0.981} \\
M3DM \cite{wang2023multimodal}   & 0.951   & 0.959  & \underline{0.912}  & 0.973   & 0.982 & \underline{0.972} & 0.967    & 0.863 \\
AST \cite{rudolph2023asymmetric} & \textbf{0.978}   & \textbf{0.983}  & 0.910  & 0.975   & 0.960 & 0.962 & \underline{0.968}    & 0.945 \\ 
UniNet \cite{wei2025uninet}  &0.920    & \underline{0.977}  &0.884   &0.975    & \textbf{0.995} & \textbf{0.996} & \textbf{0.988}    & \textbf{0.992}\\ 
\textbf{Ours}  &\underline{0.977}    & 0.969  &\textbf{0.959}   &0.962    & 0.981 & 0.953 & 0.962    & 0.978\\ \toprule
Method & Bicycle & Pizza  & Strawberry & Cabbage & Plaid & Crab  & Radish   & Mean  \\ \toprule
RGB iNet   & 0.966   & 0.826  & \underline{0.974}      & 0.975   & 0.942 & 0.961 & 0.794    & 0.920 \\
Depth iNet & 0.707   & 0.699  & 0.922      & 0.828   & 0.453 & 0.609 & 0.878    & 0.719 \\
RGB+FPFH   & 0.946   & 0.975  & 0.971      & 0.980   & 0.920 & 0.967 & \underline{0.947}    & 0.950 \\
M3DM \cite{wang2023multimodal}       & 0.961   & 0.978  & 0.930      & 0.945   & \textbf{0.976} & \underline{0.983} & \textbf{0.980}    & 0.955 \\
AST \cite{rudolph2023asymmetric}        & \underline{0.980}   & \textbf{0.980}  & 0.967      & 0.958   & 0.943 & 0.922 & 0.939    & 0.958 \\ 
UniNet \cite{wei2025uninet} &\textbf{0.989}  &\underline{0.979}  &\textbf{0.996}  &\textbf{0.997}   &\underline{0.975}  &\textbf{0.991}  &0.872   &\textbf{0.968}\\ 
\textbf{Ours} & 0.916      &\textbf{0.980}      &0.942     &\underline{0.983}     &0.929     &0.967    &\underline{0.947}      &\underline{0.960}\\ \bottomrule
\end{tabular}}
\end{table*}

\begin{table*}[ht]
\centering
\caption{{Object category-wise AUROC on PD-REAL. Results are reported per category, with Mean denoting the average AUROC across all categories. Best and second-best results are highlighted in bold and underlined, respectively. Compared with AUPRO in Tab.~\ref{tab:tab1}, AUROC better reflects image-level reliability under false-positive trade-offs.}}
\label{tab:tab2}
\scalebox{0.8}{
\begin{tabular}{ccccccccc}
\toprule
Method     & Chicken & Banana & Cookie     & Bread   & Sushi & Car   & Airplane & Train \\ \toprule
RGB iNet   & 0.653   & 0.884  & 0.615      & \textbf{0.997}   & \textbf{1.000} & \underline{0.942} & \textbf{1.000}    & \underline{0.995} \\
Depth iNet & 0.433   & 0.458  & 0.717      & 0.241   & 0.585 & 0.605 & 0.845    & 0.863 \\
RGB+FPFH   & 0.953   & 0.823  & 0.643      & 0.916   & \textbf{1.000} & 0.828 & \underline{0.982}    & 0.982 \\
M3DM \cite{wang2023multimodal}      & \underline{0.992}   & 0.646  & \underline{0.871}      & 0.543   & \underline{0.947} & 0.892 & 0.585    & 0.390 \\
AST \cite{rudolph2023asymmetric}       & \textbf{1.000}   & \textbf{0.973}  & 0.870      & 0.853   & 0.880 & 0.930 & \textbf{1.000}    & 0.967 \\ 
UniNet \cite{wei2025uninet} &0.698  &0.894  &0.247   &0.877  &\textbf{1.000}    &\textbf{1.000}   &\textbf{1.000}  &\textbf{1.000}\\ 
\textbf{Ours}  & 0.905     &\underline{0.903}       &\textbf{0.873}       &\underline{0.994}       &\textbf{1.000}      & 0.938      &0.948       &0.990\\ \toprule

Method     & Bicycle & Pizza  & Strawberry & Cabbage & Plaid & Crab  & Radish   & Mean  \\ \toprule
RGB iNet   & \textbf{1.000}   & \underline{0.995}  & \textbf{1.000}      & \textbf{1.000}   & \underline{0.855} & \textbf{1.000} & 0.937    & 0.925 \\
Depth iNet & 0.742   & 0.813  & 0.722      & 0.635   & 0.363 & 0.672 & 0.706    & 0.627 \\
RGB+FPFH   & 0.941   & \textbf{1.000}  & 0.903      & 0.930   & 0.743 & \underline{0.992} & 0.922    & 0.904 \\
M3DM \cite{wang2023multimodal}       & 0.850   & 0.990  & 0.668      & 0.793   & 0.795 & 0.850 & \underline{0.982}    & 0.786 \\
AST \cite{rudolph2023asymmetric}        & \underline{0.990}   & \textbf{1.000}  & \underline{0.998}      & 0.875   & \textbf{0.996} & 0.748 & 0.908    & \underline{0.933} \\ 
UniNet \cite{wei2025uninet} &\textbf{1.000}  &0.928   &\textbf{1.000}   &\textbf{1.000}   &0.852   &\textbf{1.000}   &0.861  &0.891\\ 
\textbf{Ours} & 0.946    & \underline{0.995}   & 0.993    &\underline{0.990}     & 0.777      &\textbf{1.000}        &\textbf{0.997}         & \textbf{0.950}\\ \bottomrule
\end{tabular}}
\end{table*}

\begin{table*}[]
\centering
\caption{{Anomaly type-wise AUPRO on PD-REAL. Six anomaly types are evaluated (dent, crack, perforation, scratch, combine-S, and combine-D). Best and second-best results are highlighted in bold and underline, respectively.}}
\label{tab:tab3}
\scalebox{0.85}{
\begin{tabular}{lcccccc}
\toprule
\multicolumn{1}{c}{Method} & Dent  & Crack & Perforation & Scratch & Combine-S & Combine-D \\ \hline
RGB iNET                   & 0.946 & 0.979 & 0.982       & 0.719   & \underline{0.968}     & 0.983     \\
Depth iNET                 & 0.626 & 0.835 & 0.863       & 0.726   & 0.560     & 0.751     \\
RGB+FPFH                   & \underline{0.976} & \underline{0.981} & \underline{0.984}       & 0.853   & 0.932     & 0.983     \\ 
M3DM \cite{wang2023multimodal}  & 0.975 & 0.978 & \underline{0.984}       & 0.758   & 0.965     & \underline{0.984}    \\ 
AST \cite{rudolph2023asymmetric}                       &0.955      &0.965      & 0.974           &\underline{0.904}    &0.853      &0.981     \\
UniNet \cite{wei2025uninet} &0.965    &\textbf{0.995}     &\textbf{0.999}       &0.827        &\textbf{0.986}       &\textbf{0.999} \\
\textbf{Ours}   & \textbf{0.977}    & 0.972      & 0.976     & \textbf{0.958}      & 0.929       & 0.983     \\
\bottomrule
\end{tabular}}
\end{table*}

\begin{table*}[ht]
\centering
\caption{{Anomaly type-wise AUROC on PD-REAL. Six anomaly types are evaluated (dent, crack, perforation, scratch, combine-S, and combine-D). Best and second-best results are highlighted in bold and underline, respectively.}}
\label{tab:tab4}
\scalebox{0.85}{
\begin{tabular}{lcccccc}
\toprule
\multicolumn{1}{c}{Method} & Dent  & Crack & Perforation & Scratch & Combine-S & Combine-D \\ \hline
RGB iNET                   & \underline{0.958} & \textbf{0.991} & \textbf{1.000}       & 0.471   & \textbf{0.848}     & \textbf{1.000}     \\
Depth iNET                 & 0.543 & 0.721 & 0.800       & 0.457   & 0.483     & 0.570     \\
RGB+FPFH                   & 0.951 & 0.959 & 0.986       & 0.535   & 0.675     & 0.993     \\ 
M3DM \cite{wang2023multimodal}                        & 0.749 & 0.837 & 0.952       & \underline{0.662}   & 0.702     & \textbf{1.000}     \\ 
AST  \cite{rudolph2023asymmetric}                      &0.916      &0.922      & \underline{0.997}           &0.637    &0.681      &\textbf{1.000} \\
UniNet \cite{wei2025uninet} &0.929       &\underline{0.988}        &\textbf{1.000}           &0.393           &\underline{0.806}         &\textbf{1.000}      \\
\textbf{Ours}   & \textbf{0.967}   & \underline{0.988}      & 0.991     & \textbf{0.830}      & 0.787       & \underline{0.998}     \\
\bottomrule
\end{tabular}}
\end{table*}

\begin{figure*}[ht]
     \centering
     \includegraphics[width=1\linewidth]{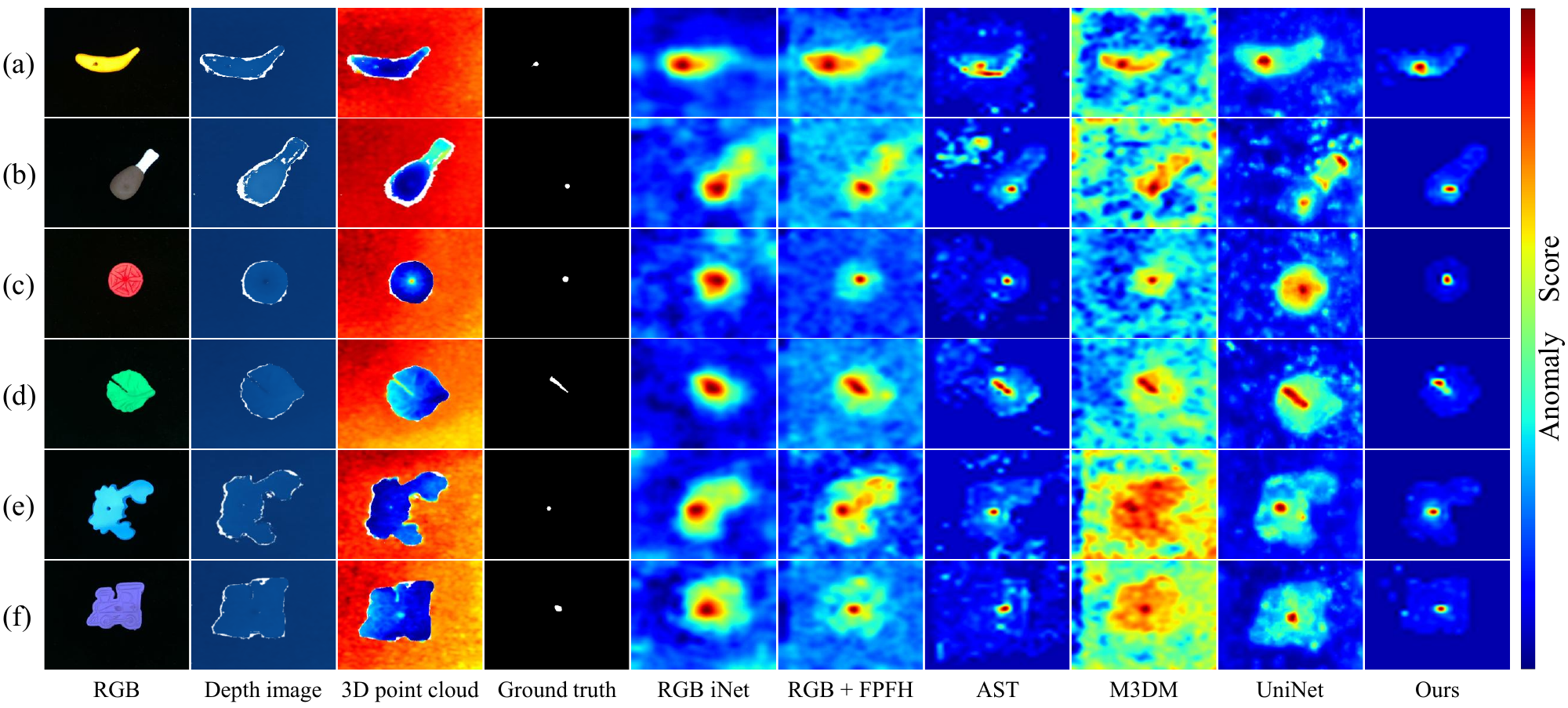 }
     \caption{{Qualitative results for anomaly localization on PD-REAL. From left to right: RGB image, depth image, 3D point cloud, ground truth, and anomaly maps from different methods (RGB iNET, Depth iNET, RGB+FPFH, AST, M3DM, UniNet, and ours).}}
     \label{fig:8}
\end{figure*}

\subsection{Comparison with State-of-the-Art Methods}
\qin{The results in Tab. \ref{tab:tab1} and Tab. \ref{tab:tab2} underscore the efficacy of our model. Our method achieves the highest AUROC among all compared methods and maintains near-optimal AUPRO scores, only marginally trailing the best-performing competitor. These results collectively validate our method's robust SOTA performance.  Minor gaps on certain categories aside, our method sustains the lowest False Positive Rate (FPR) across all state-of-the-art methods, as shown in Fig. \ref{fig:8}. In industrial inspection, FPR is a critical yet frequently underestimated metric. False positives incorrectly localize normal surface regions as defective, misdirecting operator attention toward fault-free areas. Sustained exposure to such erroneous localization triggers alarm fatigue, progressively degrading inspection reliability. More critically, as operator vigilance diminishes, genuine anomalous regions risk being overlooked entirely, allowing flawed products to bypass quality control. High FPR do more than hinder efficiency. They actively undermine the integrity of the entire detection process. Our method confronts this challenge directly. By achieving superior false-positive suppression while preserving localization accuracy, our approach ensures a level of reliability. This balance is essential for any trustworthy industrial deployment. A comprehensive analysis of anomaly detection results, organized by category and anomaly type, is presented below.}\par

\noindent	\textbf{Category-wise AD performance.} 
First, we report category-wise anomaly localization results. Considering both Table~\ref{tab:tab1} and the qualitative maps in Fig.~\ref{fig:8}, methods that fuse image and 3D point cloud cues generally provide more reliable anomaly localization than image-only settings. Although the detection performance in Fig.~\ref{fig:8}(a) is comparable, false positives increase markedly when the 2D image does not clearly indicate anomalies (Fig.~\ref{fig:8}(b,c)). In addition, for PatchCore-based methods, RGB-only input performs better than RGB+FPFH for image-level AD (Tab.~\ref{tab:tab2}, rows 2$\sim$4 and 10$\sim$12). \qin{However, for some categories (e.g., chicken) with a large proportion of visually subtle anomalies}, such as dents, 3D information is still required to improve detection performance. This analysis further confirms the importance of 3D information for AD tasks. For \cite{wang2023multimodal} and \cite{rudolph2023asymmetric}, both methods perform well on pixel-level AD (AUPRO). In particular, \cite{rudolph2023asymmetric} not only achieves higher segmentation accuracy but also significantly reduces false-positive rates (7th column in Fig.~\ref{fig:8}). A possible reason is that the asymmetric teacher--student network effectively mitigates unexpected generalization in knowledge distillation. \qin{Despite its effectiveness, \cite{rudolph2023asymmetric} still suffers from the limited receptive field of single-scale distillation, which makes it difficult to reconcile global context with local details and leads to small false positives in the detection results (7th column in Fig.~\ref{fig:8}). By contrast, our multi-scale framework effectively suppresses these false positives by balancing global and local features (10th column in Fig.~\ref{fig:8}). Through multi-scale distillation, the model captures richer feature representations while avoiding overemphasis on local textures. As a result, this hierarchical design highlights genuine anomalies while suppressing spurious responses. Furthermore, }Tab.~\ref{tab:tab2} also shows that both methods \cite{wang2023multimodal, rudolph2023asymmetric} still have room for improvement in image-level AD (AUROC), especially \cite{wang2023multimodal}. \qin{While UniNet \cite{wei2025uninet} achieves a slightly higher average AUPRO, its average AUROC drops markedly, indicating weaker image-level discrimination. More importantly, Fig.~\ref{fig:8} shows that UniNet introduces substantially more false positives, which can severely reduce practical reliability in inspection scenarios. In contrast, our method preserves competitive localization while achieving better image-level detection and much stronger false-positive suppression, yielding a more balanced and deployment-oriented performance profile. Remarkably, our approach yields a substantial improvement in accuracy compared to existing methods (Tab. \ref{tab:tab2} and Fig. \ref{fig:8}). This gain is primarily attributable to the stronger feature extraction capability enabled by multi-scale distillation, further demonstrating the effectiveness of our method. }



 \noindent	\textbf{Anomaly type-wise AD performance.}
 Next, we examine the performance of AD by anomaly type to gain more insight into PD-REAL. We notice that naively including additional 3D information does not necessarily have a positive influence for all anomaly types. For example, for the anomaly type \textit{scratch}, both detection and localization are significantly improved by using 3D and RGB (5th column in Tab. \ref{tab:tab3} and \ref{tab:tab4}). This is likely because many scratch defects induce only subtle geometric perturbations, making RGB-only cues less reliable for robust detection and localization. In contrast, for \textit{combine-S}, since the combined foreign part to the original sample is tiny, the 3D information in such a case is more like noise, thus hindering the overall image-level detection performance (6th column in Tab. \ref{tab:tab4}). We can then confirm that state-of-the-art AD methods still have a lot of room in handling 3D information appropriately for AD, which further evidences the challenge of PD-REAL.

\noindent\textbf{Comparison on MVTec 3D-AD. }\qin{ To provide an external validation beyond PD-REAL, we additionally report results on the widely-used MVTec 3D-AD dataset. Since our primary contribution is the PD-REAL dataset and a thorough benchmark on PD-REAL has been presented in Tab. \ref{tab:tab1} $\sim$~ Tab. \ref{tab:tab4}, we focus this cross-dataset evaluation on \cite{rudolph2023asymmetric}}, the second-best representative baseline on PD-REAL, to offer a concise yet informative comparison under the same evaluation metrics (AUROC for image-level detection and AUPRO for pixel-level localization). As shown in Tab. \ref{tab:tab5}, our method achieves better performance than \cite{rudolph2023asymmetric} on most categories, with a higher average AUPRO, demonstrating the improved generalization brought by multi-scale distillation.}

\begin{table}[ht]
\centering
\caption{{Comparison on the MVTec 3D-AD dataset. AUROC and AUPRO are reported for image-level detection and pixel-level localization, respectively. Mean denotes the average over all categories. Best results are highlighted in bold.}}
\label{tab:tab5}
\begin{tabular}{ccccc}
\toprule
\multirow{2}{*}{Category} & \multicolumn{2}{c}{AST} & \multicolumn{2}{c}{\textbf{Ours}} \\
                          & AUROC      & AUPRO      & AUROC       & AUPRO      \\ \hline
Begal                     & 0.936      & 0.963      & \textbf{0.940}    & \textbf{0.968}     \\
Cable\_gland          & \textbf{0.975}      & 0.945      & 0.963       & \textbf{0.947}      \\
Carrot                & \textbf{0.993}      & \textbf{0.981}      & \textbf{0.993}       & \textbf{0.981}      \\
Cookie               & 0.993      &0.938      & \textbf{1.000}       & \textbf{0.939}      \\
Dowel                & 0.984      & \textbf{0.914}      & \textbf{0.986}       & 0.913     \\
Foam                & \textbf{0.893}      & 0.897      & 0.877       & \textbf{0.928}      \\
Peach                & 0.990       & \textbf{0.981}      & \textbf{0.991}       & \textbf{0.981}      \\
Potato               & \textbf{0.998}      & \textbf{0.983}      & \textbf{0.998}       & \textbf{0.983}      \\
Rope                 & 0.947      & 0.892      & \textbf{0.955}      & \textbf{0.901}      \\
Tire                 & 0.764     & 0.939      & \textbf{0.769}       & \textbf{0.941}      \\
Mean                 & \textbf{0.947}      & 0.943      & \textbf{0.947}       & \textbf{0.948}      \\ 
\bottomrule
\end{tabular}
\end{table}

\begin{figure*}[ht]
     \centering
     \includegraphics[width=0.7\linewidth]{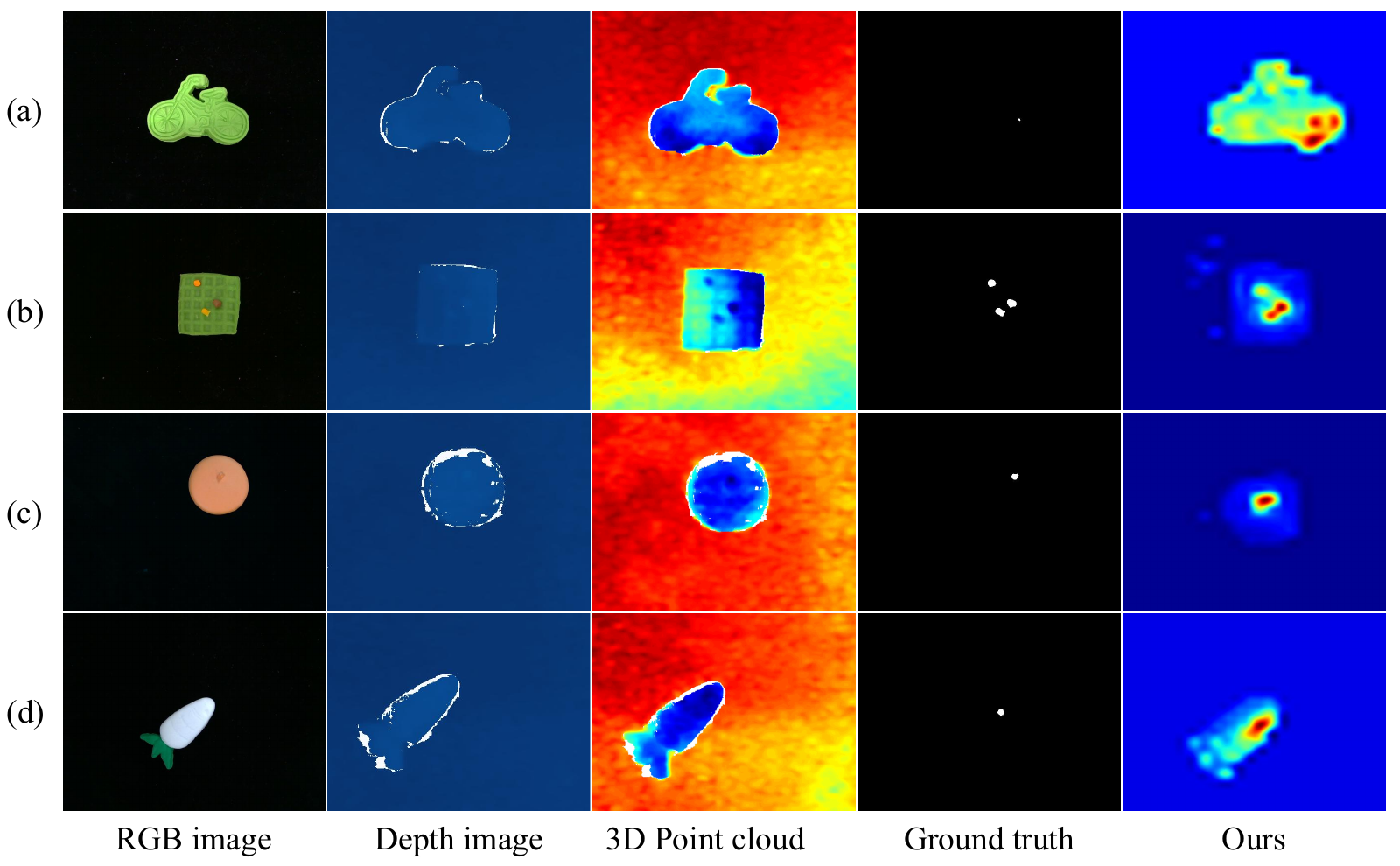 }     \caption{{Failure cases. From left to right: RGB image, depth image, 3D point cloud, ground truth, and our anomaly map. (a)--(d) show bicycle, plaid, cookie, and radish examples, respectively. }} 
     \label{fig:9}
\end{figure*}

\subsection{Ablation Studies}
To give a deeper understanding of our method, we here analyze the influence regarding the different distillation scales, as shown in Tab. \ref{tab:tab6}.

\begin{table*}[h!]
\centering
\caption{{Average AUROC and AUPRO over all categories with different distillation positions. $\tau_1$,$\tau_2$, and $\tau_3$ denote single-scale distillation at local-, mid-, and global-level feature stages, while combinations indicate multi-scale distillation. The best value in each row is highlighted in bold.}}
\label{tab:tab6}
\begin{tabular}{cccccccc}
\toprule
Num of distillation positions & $\tau_1$ &$\tau_2$ &$\tau_3$ &$\tau_1+\tau_2$ &$\tau_1+\tau_3$ &$ \tau_2+\tau_3$ & $\tau_1+\tau_2+\tau_3$ \\ \hline
Avg. AUROC                    & 0.933 & 0.941 & 0.940 & 0.942 & 0.944 & 0.942 & \textbf{0.950}    \\
Avg. AUPRO                    & 0.958 & 0.959 & 0.958 & \textbf{0.960} & \textbf{0.960} & 0.959 & \textbf{0.960}    \\ 
\bottomrule
\end{tabular}
\end{table*}

As the number of distillation positions increases from single-scale to multi-scale settings, AUROC consistently improves. AUPRO also improves but quickly saturates in multi-scale configurations ($\tau_1+\tau_2$, $\tau_1+\tau_3$, and $\tau_1+\tau_2+\tau_3$). Overall, the full three-scale setting ($\tau_1+\tau_2+\tau_3$) provides the most robust and balanced performance for both detection and localization.

\subsection{Failure Cases}

Fig.~\ref{fig:9} summarizes representative failure cases of our method. In (a), the bicycle surface contains many recessed grids rather than a flat texture, which makes it difficult to learn reliable 3D priors of normal samples. When a combined object has a similar color to the bicycle and is placed inside a recess, its 3D protrusion is not salient, so the fused features provide a weak 3D cue. The defect can still be detected, but false positives appear around neighboring grids.

In (b), the plaid sample further illustrates this effect. If the foreign object sits inside a grid, its 3D signal is weak, although it can be detected due to color contrast, its anomaly score is not the strongest. In contrast, objects placed on top of the grid have more pronounced 3D geometry, leading to higher anomaly scores and more accurate localization.

In (c), the cookie sample shows that shadows and color differences near the combined object introduce false positives. Nevertheless, because the 3D cue is less sensitive to illumination, the true defect still receives the highest anomaly score, indicating the robustness brought by depth information under lighting interference.

In (d), the radish defect is extremely small and close in color to the surface, making RGB-only detection unreliable. With fused features, our method still assigns the highest score to the dent location, but produces false positives nearby. This is because the dent is subtle and normal radish surfaces already contain small pits, which complicates the modeling of normal geometry and introduces noise.

Overall, these failure cases highlight both the importance and the challenge of 3D information for anomaly detection: depth cues improve robustness to illumination and help distinguish true defects, but structured surfaces and subtle geometry can mask anomalies and trigger false positives.

\noindent\textbf{Limitations and future work.}
Despite the overall gains, several limitations remain. Structured textures (e.g., plaid or bicycle surfaces with repetitive grids) introduce normal geometric variations that can be confused with defects. Illumination changes and cast shadows may also bias RGB cues and create spurious responses. Moreover, extremely small defects are still challenging because their geometric signatures are weak and can be overwhelmed by normal micro-variations.

\section{Conclusion}
In this paper, we proposed a novel Play-Doh made dataset named PD-REAL for unsupervised 3D anomaly detection, which provides over 3,500 high-precision RGB images and corresponding depth information with six types of anomalies and 15 object categories. Compared to previous 3D AD datasets, our data collection pipeline is substantially cheaper, easy to conduct, and easily expandable. In addition, \qin{we proposed a multi-scale teacher--student framework for multimodal 3D anomaly detection, by distilling hierarchical representations from the teacher network, the student network can acquire both global and local features. Furthermore, we investigated the performance of our method and state-of-the-art AD approaches on PD-REAL, and demonstrated the superiority of our method, as well as the importance and challenge of using 3D information for AD across different object categories and anomaly types. We hope that our dataset can further stimulate the exploration of better solutions to 3D AD in the future.}

\section{Acknowledgement}
This work is supported by Ningbo Philosophy and Social Science Laboratory Project Grant Number SYS-16, JSPS KAKENHI Grant Number JP20K19568, Philosophy and Social Science Planning Cross-disciplinary Key Support Subjects of Zhejiang Province Grant Number 22JCXK08Z, and Talent Recruitment Fund of Ningbotech University Grant Number 20250327Z0090. We also acknowledge China Scholarship Council (CSC) for funding the first author under Grant Number 202108330097.

\bibliographystyle{cas-model2-names}
\bibliography{reference}

\end{document}